\documentclass[twoside]{article}

\usepackage[accepted]{aistats2020}
\usepackage{amsmath}%
\usepackage{amsfonts}%
\usepackage{amssymb}%
\usepackage{amsmath}
\usepackage{mathtools}
\usepackage{bbm}
\usepackage{graphicx}
\providecommand{\U}[1]{\protect\rule{.1in}{.1in}}
\RequirePackage[OT1]{fontenc}
\RequirePackage{amsthm,amsmath,natbib}
\usepackage{xr}
\RequirePackage{hyperref}
\usepackage{booktabs}
\usepackage{subfigure}
\usepackage{color}
\usepackage[utf8]{inputenc}

\DeclareMathOperator*{\Poisson}{Poisson}
\DeclareMathOperator*{\tExp}{tExp}

\newtheorem{proposition}{Proposition}
\newcommand{\1}[1]{\mathbbm{1}_{#1}}

\newcommand{\obsim}[1]{\overset{#1}{\boldsymbol\sim}}
\newcommand{\obasymp}[1]{\overset{#1}{\boldsymbol\asymp}}

\begin{document}

% If your paper is accepted and the title of your paper is very long,
% the style will print as headings an error message. Use the following
% command to supply a shorter title of your paper so that it can be
% used as headings.
%
\runningtitle{Non-exchangeable feature allocations models with sublinar features' growth}

% If your paper is accepted and the number of authors is large, the
% style will print as headings an error message. Use the following
% command to supply a shorter version of the authors names so that
% they can be used as headings (for example, use only the surnames)
%
%\runningauthor{Surname 1, Surname 2, Surname 3, ...., Surname n}

\twocolumn[

\aistatstitle{Non-exchangeable feature allocation models\\ with sublinear growth of the feature sizes}

\aistatsauthor{Giuseppe Di Benedetto \And  Fran\c cois Caron \And Yee Whye Teh}
\aistatsaddress{University of Oxford \And University of Oxford \And University of Oxford \\ DeepMind}
]

\begin{abstract}
Feature allocation models are popular models used in different applications such as unsupervised learning or network modeling. In particular, the Indian buffet process is a flexible and simple one-parameter feature allocation model where the number of features grows unboundedly with the number of objects. The Indian buffet process, like most feature allocation models, satisfies a  symmetry property of exchangeability: the distribution is invariant under permutation of the objects. While this property is desirable in some cases, it has some strong implications. Importantly, the number of objects sharing a particular feature grows linearly with the number of objects. In this article, we describe a class of non-exchangeable feature allocation models where the number of objects sharing a given feature grows sublinearly, where the rate can be controlled by a tuning parameter. We derive the asymptotic properties of the model, and show that such model provides a better fit and better predictive performances on various datasets.
\end{abstract}

\section{Introduction}

Feature allocation models are probabilistic models over multisets~\citep{Broderick2013}, which represent the allocation of a set of objects to a (potentially unbounded) set of features. Contrary to models on partitions, where each object is assigned to a single group, these models allow each object to be allocated more than one feature. For example, the objects may be movies, and the features correspond to the actors performing in that movie. Movies have different number of actors, and actors participate to a different number of movies. Informally, feature allocations models can be interpreted as distributions over sparse binary matrices whose rows represent the objects and whose columns represent the features; non-zero entries indicate the allocation of features to objects. Feature allocation models have been used in various applications, including  topic modeling~\citep{Williamson2010}, image analysis~\citep{Zhou2011a}, network modeling~\citep{Palla2012,Cai2016} or inference in tumor heterogeneity~\citep{Lee2015,Xu2015}.

A classical assumption is that of exchangeability: the feature allocation model is invariant over permutations of the objects. Taking the interpretation as a binary matrix, the matrix is invariant over permutations of its rows. The most remarkable example of an exchangeable feature allocation is the Indian buffet process (IBP)  \citep{Griffiths2005InfiniteLF,Thibaux2007,Griffiths2011TheIB} which has a simple and intuitive generative model. The model allows the number of features $K_n$ to grow unboundedly with the number of objects $n$ at a logarithmic rate. The IBP admits a three-parameter generalisation, the stable IBP~\citep{Teh2009}, which is also exchangeable. For some values of its parameters, the stable IBP can capture a power-law behavior, where the number of features grows at a rate of $n^\sigma$ for some $\sigma\in(0,1)$.

While the exchangeability assumption is reasonable for many applications and has computational advantages, it may not be adequate in some cases. In particular, assuming exchangeability implies that, out of $n$ objects, the number $m_{n,j}\leq n$ of objects  having a particular feature $j$ (called feature's size) scales linearly with the number of objects $n$ (see Figure 4(b) for an illustration). Such assumption may be undesirable. For instance, even if she's a prolific actress, one does not expect the filmography of Meryl Streep to scale linearly with the overall number of movies released.

The objective of this article is to present a model that can have a sublinear growth of the size of the features, while retaining the properties of the (stable) Indian buffet in terms of overall growth of the number of features and power-law properties. The article is organised as follows. Section \ref{sec:background} provides some background on feature allocation models and the IBP. Our non-exchangeable model is presented in Section \ref{model} and its asymptotic properties given in Section~\ref{sec:asymptotics}. In Section \ref{sec:inference} we derive a Gibbs sampler for posterior inference. Experimental results are presented in Section \ref{experiments}. Related approaches are discussed in Section \ref{sec:discussion}.

\section{Background}
\label{sec:background}
\subsection{Feature allocation models and the Indian buffet process}

A feature allocation \citep{Broderick2012} $f_n=\{A_{n,1},\ldots,A_{n,K_n}\}$ is a multiset of a set of objects $\{1,\ldots,n\}$ such that $A_{n,k}$, $k=1,\ldots,K_n$ are (possibly overlapping) non-empty subsets of $\{1,\ldots,n\}$; $A_{n,k}$ represents the set of objects having feature $k$ and $K_n$ is the number of different features shared by the $n$ objects.  For example, $f_4=\{\{1,3\},\{1,3\},\{3,4\},\{4\}, \{4\}\}$ indicates that object 1 has features 1 and 2, object 2 has no feature, object 3 has features 1, 2, 3 and object 4 has features 3, 4 and 5. Note that the labelling of the features as 1 to 5 is arbitrary.

A feature allocation model is a distribution over a growing family of random feature allocations $(f_n)_{n=1,2,\ldots}$. The most popular feature allocation model is the Indian buffet process, where $f_n$ has distribution proportional to
\begin{equation}
\Pr(f_n)\propto \eta^{K_n} \prod_{k=1}^{K_n} \frac{(\widetilde m_{n,k}-1)!(n-\widetilde m_{n,k})!}{n!}
\label{eq:featureallocationIBP}
\end{equation}
where $\widetilde m_{n,k}$ is the number of objects having feature $k$, for $k=1,\ldots,K_n$ and $\eta>0$ is a tuning parameter. The IBP admits a three-parameter generalisation, called stable IBP~\citep{Teh2009}, where\footnote{Note that we use a slightly different parameterisation compared to that of \cite{Teh2009}.}
\begin{equation}
\Pr(f_n)\propto \frac{\eta^{K_n}}{\Gamma(1-\sigma)^{K_n}}\prod_{k=1}^{K_n} \frac{\Gamma(\widetilde m_{n,k}-\sigma)\Gamma(n-\widetilde m_{n,k}+\zeta)}{\Gamma(n+\zeta-\sigma)}
\label{eq:featureallocationIBP}
\end{equation}
with $\eta>0$, $\sigma\in(-\infty, 1)$ and $\zeta>0$. It reduces to the one-parameter IBP when $\zeta=1$ and $\sigma=0$. When $\sigma>0$, the model exhibits power-law properties.

A convenient way to encode a feature allocation model is via a collection of atomic random measures $(Z_1,\ldots,Z_n)$ on some space (here $\mathbb R_+$) where for $i=1,\ldots,n$,
\begin{equation}
Z_i=\sum_{j\geq 1}z_{ij}\delta_{\theta_j}
\label{eq:pointmeasure}
\end{equation}
 with $z_{ij}=1$ if object $i$ has feature $j$ and $(\theta_j)_{j\geq 1}$ are continuous random variables on $\mathbb R_+$ whose distribution is irrelevant here. Note that in this notation there is no particular ordering of the features, and we now use the index $j$ instead of $k$ to emphasize this difference.

\subsection{Completely random measures}
\label{sec:CRM}
A homogeneous completely random measure (CRM) \citep{Kingman1967,Lijoi2010} on $\mathbb R_+=[0,\infty)$ is an almost surely discrete random measure %defined as
\begin{equation}
B=\sum_{j\geq 1}\omega_j\delta_{\theta_j}
\label{eq:CRM}
\end{equation}
where $\{(\omega_j,\theta_j)_{j\geq 1}\}$ are points of a Poisson point process on $(0,\infty)\times \mathbb R_+$ with mean measure
%$\nu$ that factorizes as
$\nu(d\omega,d\theta) = \rho(\omega)d\omega\alpha(\theta)d\theta$ where $\rho$ and $\alpha$ satisfy
$$
\int_0^\infty (1-e^{-\omega})\rho(\omega)d\omega<\infty~~\text{and }\int_A \alpha(\theta)d\theta<\infty
$$
for any bounded set $A\subset \mathbb R_+$. The above condition ensures that $B(A)<\infty$ almost surely. The CRM is said to be infinite-activity if $\int_0^\infty \rho(\omega)d\omega=\infty$. In this case, the measure $B$ has a countably infinite support on any non-empty interval $A\subset \mathbb R_+$.

The (stable) Indian buffet process admits the following hierarchical construction via CRMs~\citep{Thibaux2007,Teh2009}. Let $\widetilde B=\sum_j \pi_j\delta_{\theta_j}$ be a homogeneous CRM with
\begin{align}
\rho(\pi)=\frac{\eta}{\Gamma(1-\sigma)} \pi^{-1-\sigma} (1-\pi)^{\zeta-1} ~\1{\pi\in(0,1)}
\end{align}
and $\alpha(\theta)=\1{\theta\leq 1}$. The parameter $\pi_j$ can be interpreted as the popularity of feature $j$. For $i=1,\ldots,n$ define the atomic measure $Z_i$ as in Equation~\eqref{eq:pointmeasure} with  $$z_{ij}\mid \pi_j\sim \text{Ber}(\pi_j)$$ for $j\geq 1$,
where $\text{Ber}(\pi)$ denotes the Bernoulli distribution with parameter $\pi\in[0,1]$.

\begin{figure*}[h!]
  \centering
  \subfigure[Beta-Bernoulli process]{\includegraphics[width=.3\textwidth]{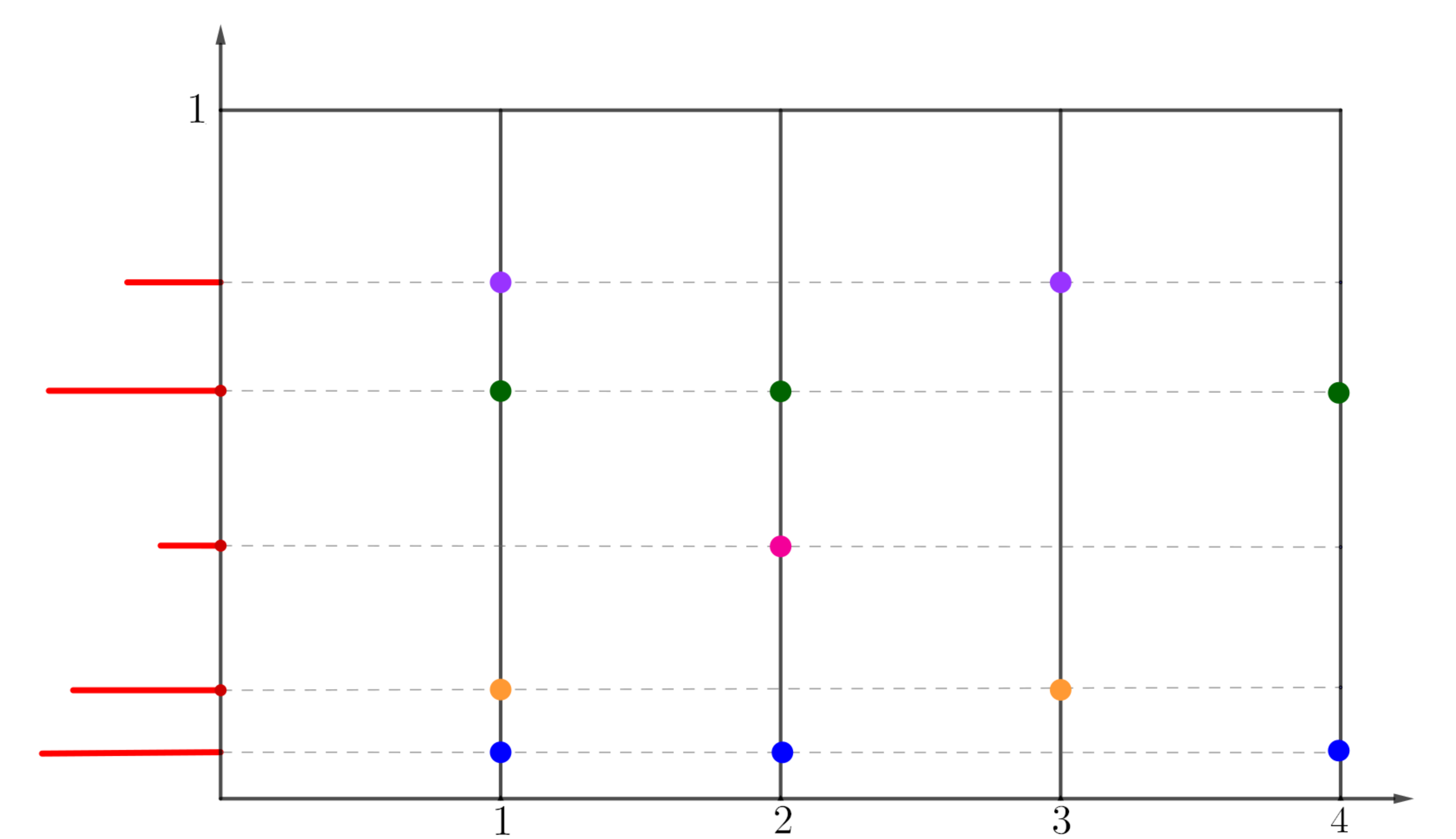}}
   \subfigure[Continuous time IBP]{\includegraphics[width=.3\textwidth]{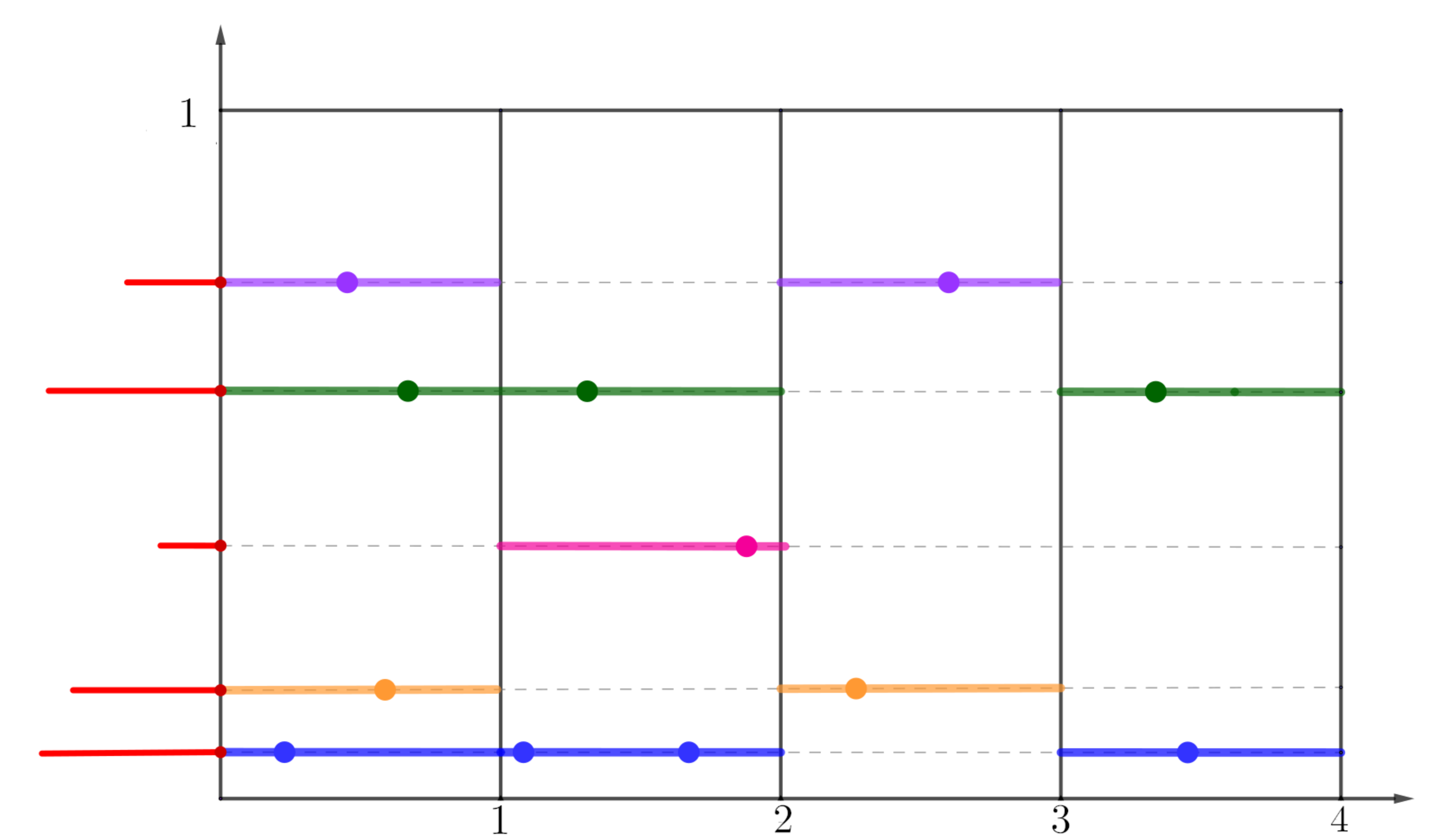}}
   \subfigure[Non-exchangeable point process]{\includegraphics[width=.3\textwidth]{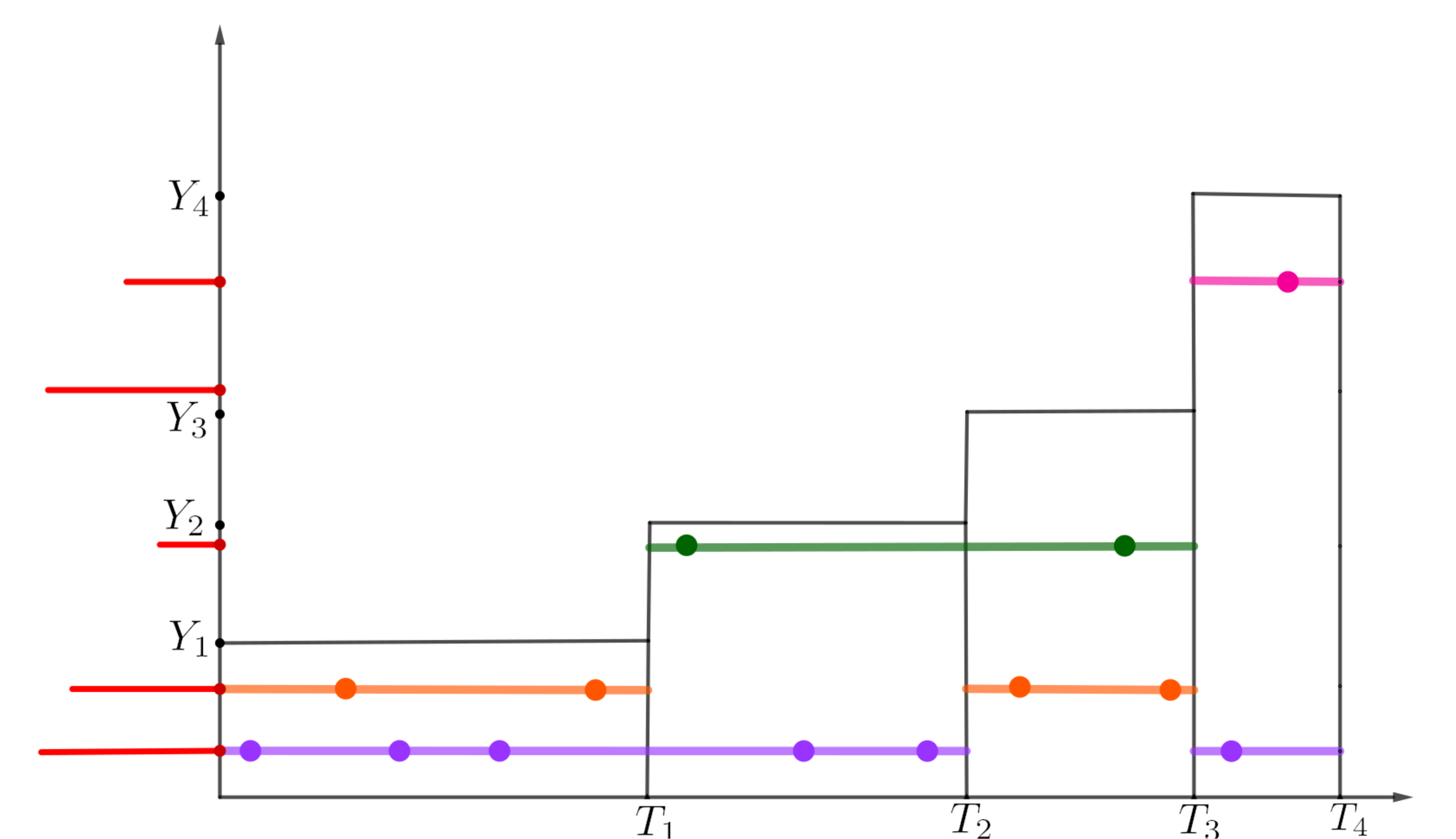}}
  \caption{Point process defining the feature models: red sticks represent weights of the CRM, points and bars of the same colour refer to the same feature. (a) Beta-Bernoulli process; (b) Point process construction of the IBP: each rectangle of basis $[T_{i-1}, T_i]$ represents the $i-th$ object whose features are depicted by the coloured bars, which are present if at least one of the corresponding points falls in the rectangle; (c) Non-exchangeable point process.}
  \label{fig:cartoon_plots}
\end{figure*}

\subsection{An alternative construction for the Indian buffet process}

We present here an equivalent construction for the IBP. It relies on the introduction of latent Poisson processes, similar to the construction proposed by \cite{DiBenedetto2017} for non-exchangeable random partitions, adapted to feature models. This approach will give the intuition for the generalisation of the IBP introduced in the next section.

Using the change of variable $\omega_j=-\log(1-\pi_j)\in(0,\infty)$, the random measure $B=\sum_j \omega_j\delta_{\theta_j}$ is itself a CRM with $\alpha(\theta)=\1{\theta\leq 1}$ and
\begin{align}
\rho(\omega)=\frac{\eta}{\Gamma(1-\sigma)} e^{-\zeta\omega}(1-e^{-\omega})^{-1-\sigma}.\label{eq:transformedbetameasure}
\end{align}
We call \eqref{eq:transformedbetameasure} a transformed stable beta (TSB) L\'evy measure, and the associated random measure $B$ a TSB process (TSBP). Consider for each $j$ a homogeneous Poisson process $N_j(t)$ on $\mathbb R_+$ with rate $\omega_j$. Let  $z_{ij}=\1{N_j(i)-N_j(i-1)>0}$ be a binary variable indicating if there is any event in the time interval $[i-1,i)$. Then
\begin{align*}
\Pr(z_{ij}=1\mid \omega_j)&=\Pr(N_j(i)-N_j(i-1)>0\mid \omega_j)\\&=1-e^{-\omega_j}=\pi_j.
\end{align*}
This construction is illustrated in Figure~\ref{fig:cartoon_plots}.

\section{Non-exchangeable feature allocation model}
\label{model}
Let $B =\sum_{j\ge1} \omega_j \delta_{\theta_j}$
be a homogeneous completely random measure on $\mathbb R_+$. While the model can be defined for a general CRM, we focus here on the case where $\alpha(\theta)=1$ and  $\rho$ is either the transformed stable beta measure \eqref{eq:transformedbetameasure}, or the generalised gamma (GG) measure~\citep{Hougaard1986}, given by
\begin{equation}
\rho(\omega)=\frac{\eta}{\Gamma(1-\sigma)}\omega^{-1-\sigma}e^{-\zeta\omega}\1{\omega>0}\label{eq:ggpmeasure}
\end{equation}
where $\eta>0$, $\sigma\in(-\infty,1)$, $\zeta>0$. As we will show in Section~\ref{sec:asymptotics}, both models lead to the same asymptotic behavior. The GG measure has however a conjugate form that makes it more amenable to posterior inference, as detailed in Section~\ref{sec:inference}. A CRM with mean measure \eqref{eq:ggpmeasure} will be called a Generalised Gamma Process (GGP).

Inspired by the work on random partition models by \cite{DiBenedetto2017}, let  $(Y_n)_{n\geq 1}$ and $(T_n)_{n\geq 1}$ be two increasing sequences of positive reals defined as
\begin{align}
T_n =n^{\frac{1}{\xi+1}} ,\qquad Y_n = n^{\frac{\xi}{\xi+1}}
\end{align}
where $\xi\geq 0$ is a tuning parameter. Define the sequence $(\Delta_n)_{n\geq 1}$ by $\Delta_n:=T_n-T_{n-1}$. The feature allocation of an object $i$ is represented by a random measure $Z_i$ on $\mathbb R_+$ as in Equation~\eqref{eq:pointmeasure} where
\begin{align} z_{ij}\mid \omega_j,\theta_j \sim\text{Ber}\left(1-e^{-\omega_j\Delta_i\1{\theta_j\leq  Y_i}}\right).\label{eq:zijnonexch}
\end{align}
Note that $z_{ij}=0$ a.s. if $\theta_j>Y_i$.

The model admits the following construction using a latent Poisson process. For each $j$, let $(N_j(t))_{t>0}$ be a homogeneous Poisson process with rate $\omega_j$. Then the binary variable $z_{ij}=\1{N_j(T_i)-N_j(T_{i-1})>0}\1{\theta_i\leq Y_i}$ has distribution \eqref{eq:zijnonexch}. See the illustration in Figure~\ref{fig:cartoon_plots}(c).

Note that if one sets $\xi=0$, we have $\Delta_i=Y_i=1$. The distribution in the right-handside of Equation~\eqref{eq:zijnonexch} does not depend on $i$ and the associated feature allocation model is therefore exchangeable.  If additionally we use the mean measure $\rho$ as in Equation~\eqref{eq:transformedbetameasure}, we recover the three-parameter IBP as a special case.

The model is parameterised by the four parameters $\eta$, $\sigma$, $\zeta$ and $\xi$. In the next section we show how these parameters tune the asymptotic properties of the model. The critical parameter is the parameter $\xi$ and we show that for $\xi>0$, the features' sizes grow sublinearly with $n$, at a rate controlled by this parameter.

\section{Asymptotic Properties}\label{sec:asymptotics}

\subsection{Notations}

In this section we use the following notations for asymptotics. $a_n\obsim{n\to\infty} b_n$ means that $\lim_{n\to\infty} \frac{a_n}{b_n}=1$ and $a_n\obasymp{n\to\infty} b_n$ means $\limsup_{n\to\infty} \frac{a_n}{b_n}<\infty$ and  $\limsup_{n\to\infty} \frac{b_n}{a_n}<\infty$ (that is, $a_n$ is of the same order as $b_n$).
Let $$Z_i(\mathbb R_+)=\sum_{j\geq 1} z_{ij}$$ be the number of features for object $i$. For each feature $j=1,2,\ldots$, let us define its size as $$m_{n,j}=\sum_{i=1}^n z_{ij}$$
i.e.  the number of objects having that feature, and let $$m_n=\sum_{j\geq 1}m_{n,j}=\sum_{i=1}^n\sum_{j\geq 1} z_{ij}$$ be the total number of features allocated to the $n$ objects. Denote $$K_n=\sum_j \1{m_{n,j}>0}$$ the number of unique features, and $$K_{n,r}=\sum_j \1{m_{n,j}=r}$$ the number of unique features allocated to $r$ objects, where $r\geq 1$. Note that $\sum_r K_{n,r}=K_n$.

We will use the following notation for the Laplace exponent and the tilted moments
\begin{align*}
\psi(t)&=\int_0^\infty \left \{1-e^{-\omega t} \right \}\rho(\omega)d\omega\\
\kappa(m,u)&=\int_0^\infty\omega^me^{-u\omega} \rho(\omega)d\omega.
\end{align*}
for any integer $m\geq 1$ and any $u>0$. $\kappa(m,0)$ correspond to the $m$-th moment of the measure $\rho$. For the GG measure, we have $\psi(t)=\frac{\eta}{\sigma}((t+\zeta)^\sigma-\zeta^\sigma)$ and $\kappa(m,u)=\eta\frac{\Gamma(m-\sigma)}{\Gamma(1-\sigma)}(u+\zeta)^{\sigma-m}$. For the TSB, we have $\psi(1)=\frac{\eta\Gamma(\zeta)}{\Gamma(\zeta+1-\sigma)}$, but there is no analytical expression for $\psi(t)$ and $\kappa(m,u)$.

\subsection{Asymptotic properties when $\xi=0$}
\label{sec:asymptoticsIBP}

We first recall here, for comparison, the asymptotic properties of our model for $\xi=0$ when $B$ is a GGP or a TSBP. As mentioned in Section~\ref{model}, the model is exchangeable in that case and if $B$ is the TSBP, it reduces to the three parameter IBP, whose asymptotic properties are well known~\citep{Griffiths2005InfiniteLF,Thibaux2007,Teh2009,Broderick2012}. Asymptotics when $B$ is the GGP model are similar. First, for the number of features of object $i$,
\begin{align}
Z_i(\mathbb R_+)\sim \text{Poisson}\left (\psi(1) \right )
\end{align}
where
$\psi(1)=\frac{\eta\Gamma(\zeta)}{\Gamma(\zeta+1-\sigma)}$ for the TSBP and $\psi(1)=\frac{\eta}{\sigma}((\zeta+1)^\sigma-\zeta^\sigma)$ for the GGP.

Given the CRM $B$, we have almost surely
\begin{align}
m_{n,j}&\obsim{n \to\infty} n (1-e^{-\omega_j})\\
m_n&\obsim{n \to\infty} n \sum_{j\geq 1}(1-e^{-\omega_j}).
\end{align}
Hence both the features' sizes and the total number of features grow linearly with $n$, as a consequence of the exchangeability assumptions.
For the number of unique features, we have almost surely
\begin{equation}
K_n\obsim{n\to\infty} \left \{
\begin{array}{ll}
  \eta\log(n) & \sigma=0 \\
  \frac{\eta}{\sigma}n^\sigma & \sigma\in(0,1)
\end{array}
 \right . .
\end{equation}
Finally we have, for the proportions of features allocated to $r$ objects for $\sigma\in(0,1)$
\begin{equation}
\frac{K_{n,r}}{K_n}\rightarrow\frac{\sigma\Gamma(r-\sigma)}{r!\Gamma(1-\sigma)}\label{eq:asymptoticsKnrIBP}
\end{equation}
almost surely as $n$ tends to infinity, and $K_{n,r}/K_n\rightarrow 0$ almost surely otherwise for all $r\geq 1$. Equation \eqref{eq:asymptoticsKnrIBP} corresponds to a power-law behaviour as $\Gamma(r-\sigma)/r!\simeq r^{-(1+\sigma)}$ for large $r$.

\subsection{Asymptotic properties when $\xi>0$}

\begin{figure*}[!htb]
  \centering
  \subfigure[IBP]{\includegraphics[width=.24\textwidth]{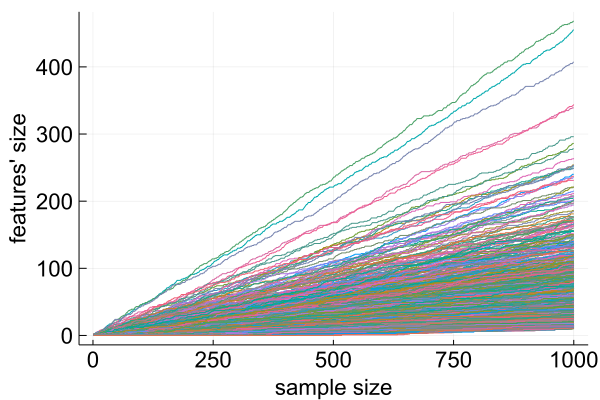}}
  \subfigure[$\xi=0.25$]{\includegraphics[width=.24\textwidth]{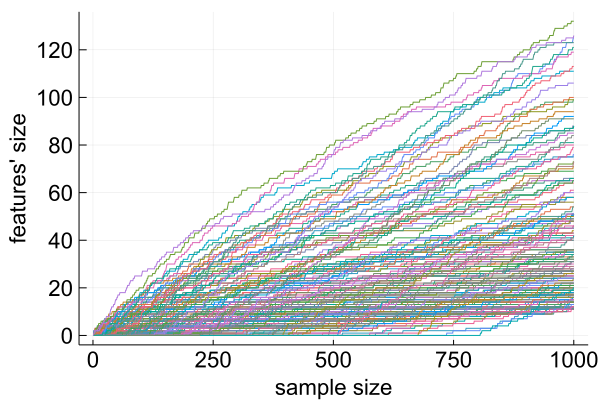}}
   \subfigure[$\xi=1$]{\includegraphics[width=.24\textwidth]{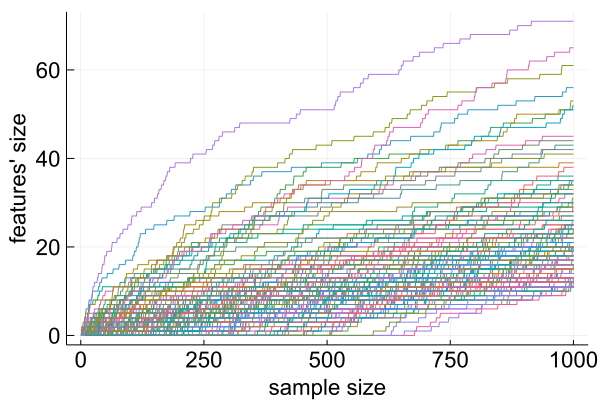}}
    \subfigure[$\xi=2$]{\includegraphics[width=.24\textwidth]{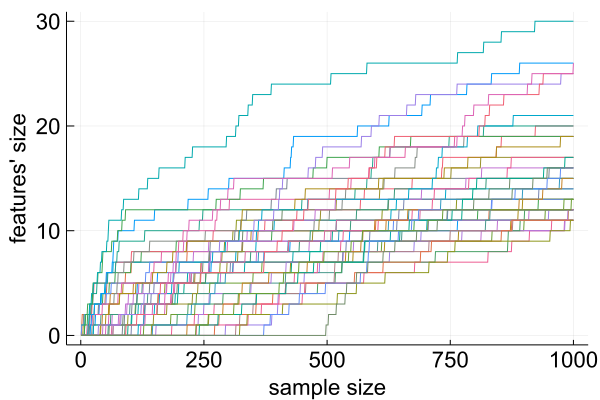}}
  \caption{Evolution of the features' sizes with the sample size for different synthetic datasets: (a) IBP; (b-d) Non-exchangeable model for $\xi=0.25, 1, 2$. The feature size growth is (a) $n$, (b) $n^{0.8}$, (c) $n^{1/2}$ and (d) $n^{1/3}$.}
  \label{fig:featuresize_synt}
\end{figure*}

When $\xi>0$, the model is non-exchangeable. In this case, most of the properties of the exchangeable case are retained, such as the Poisson number of features per object (see Proposition \ref{prop:nbfeatureobject} below), the linear growth of the total number of features
%for the $n$ objects
(Proposition \ref{prop:tot_features}) and the power-law behaviour, solely controlled by the parameter $\sigma$ (Proposition \ref{prop:powerlaw}). The number of unique features $K_n$ still grows sublinearly, but at a rate now controlled by both $\sigma$ and $\xi$ (Proposition \ref{prop:unique_features}). The key difference is that, as shown in Proposition~\ref{prop:feature_size}, the features' sizes grow at a rate $n^{1/(1+\xi)}$, which is sublinear for $\xi>0$.

Propositions  \ref{prop:nbfeatureobject}, \ref{prop:tot_features} and \ref{prop:feature_size} are valid for any choice of L\'evy measure $\rho$. Propositions \ref{prop:unique_features} and \ref{prop:powerlaw} hold for any L\'evy measure $\rho$ such that
\begin{align}
\psi(t)\obsim{t\to\infty}
\left\{
  \begin{array}{ll}
    \eta\log(t) & \sigma=0 \\
    \frac{\eta}{\sigma} t^\sigma & \sigma\in(0,1) \\
  \end{array}
\right .
.
\label{eq:asymptoticspsi}
\end{align}
Equation~\eqref{eq:asymptoticspsi} holds in particular for the GGP and the TSBP. The proofs are given in the Supplementary Material.
%Supplementary Material.

%\paragraph{Number of features per object and total number of features. }
The first proposition shows that the number of features per object is Poisson distributed, with a rate converging to a constant.
\begin{proposition}[number of features per object]
Consider the model of Section \ref{model} with a generic $\rho$ and $\xi>0$. For each object $n$, we have
\begin{align}
Z_n(\mathbb R_+)\sim \Poisson\left (\mathbb E \left [Z_n(\mathbb R_+)\right ] \right )
\end{align}
where
\begin{align*}
\mathbb E \left [Z_n(\mathbb R_+)\right ]=Y_n\psi(\Delta_n)\rightarrow (1+\xi)^{-1}\kappa(1,0),
\end{align*}%
with
$\kappa(1,0)=\eta\zeta^{\sigma-1}$ for the GGP.
\label{prop:nbfeatureobject}
\end{proposition}

The next result is on the asymptotic behaviour of the total number of features observed in the first $n$ objects.
\begin{proposition}[total number of features]
  Consider the model of Section \ref{model} with a generic $\rho$ and $\xi>0$. The total number of features observed in the first $n$ objects satisfies
  $$
  m_n \obsim{n\to\infty} (1+\xi)^{-1}\kappa(1,0)\, n.
  $$
  \label{prop:tot_features}
\end{proposition}
The following proposition describes the growth of the features' sizes $m_{n,j}$ with respect to $n$.
\begin{proposition}[number of objects per feature]\label{prop:feature_size}
  Consider the model of Section \ref{model} with a generic L\'evy measure $\rho$ and $\xi\geq 0$. For each $j\ge 1$ and conditionally on the CRM $B$, we have that, almost surely,
  $$
  m_{n,j} \obsim{n\to\infty} \omega_j T_n = \omega_j n^{1/(1+\xi)}.
  $$
\end{proposition}
The features' sizes therefore grow linearly if $\xi=0$, and sublinearly when $\xi>0$, with a rate decreasing at $\xi$ increases. This is illustrated in Figure~\ref{fig:featuresize_synt}.

\begin{proposition}[number of unique features]\label{prop:unique_features}
  Consider the model of Section \ref{model} with $\xi\geq 0$ and a L\'evy measure $\rho$ whose Laplace exponent $\psi$ satisfies Equation~\eqref{eq:asymptoticspsi}. The number of unique features $K_n$ is such that
\begin{equation}
K_n\obsim{n\to\infty}
\left\{
  \begin{array}{ll}
    \eta\,\,n^{\frac{\xi}{\xi+1}}\log(n) & \sigma=0 \\
    \eta\,\frac{\Gamma(\xi+1)\Gamma(\sigma)}{\Gamma(\sigma+\xi+1)}\,n^{\frac{\xi+\sigma}{\xi+1}} & \sigma\in(0,1) \\
  \end{array}
\right .
.
\label{eq:asymptoticsKn}
\end{equation}
\end{proposition}

\begin{figure}[t]
  \centering
  \includegraphics[width=8cm, height=4cm]{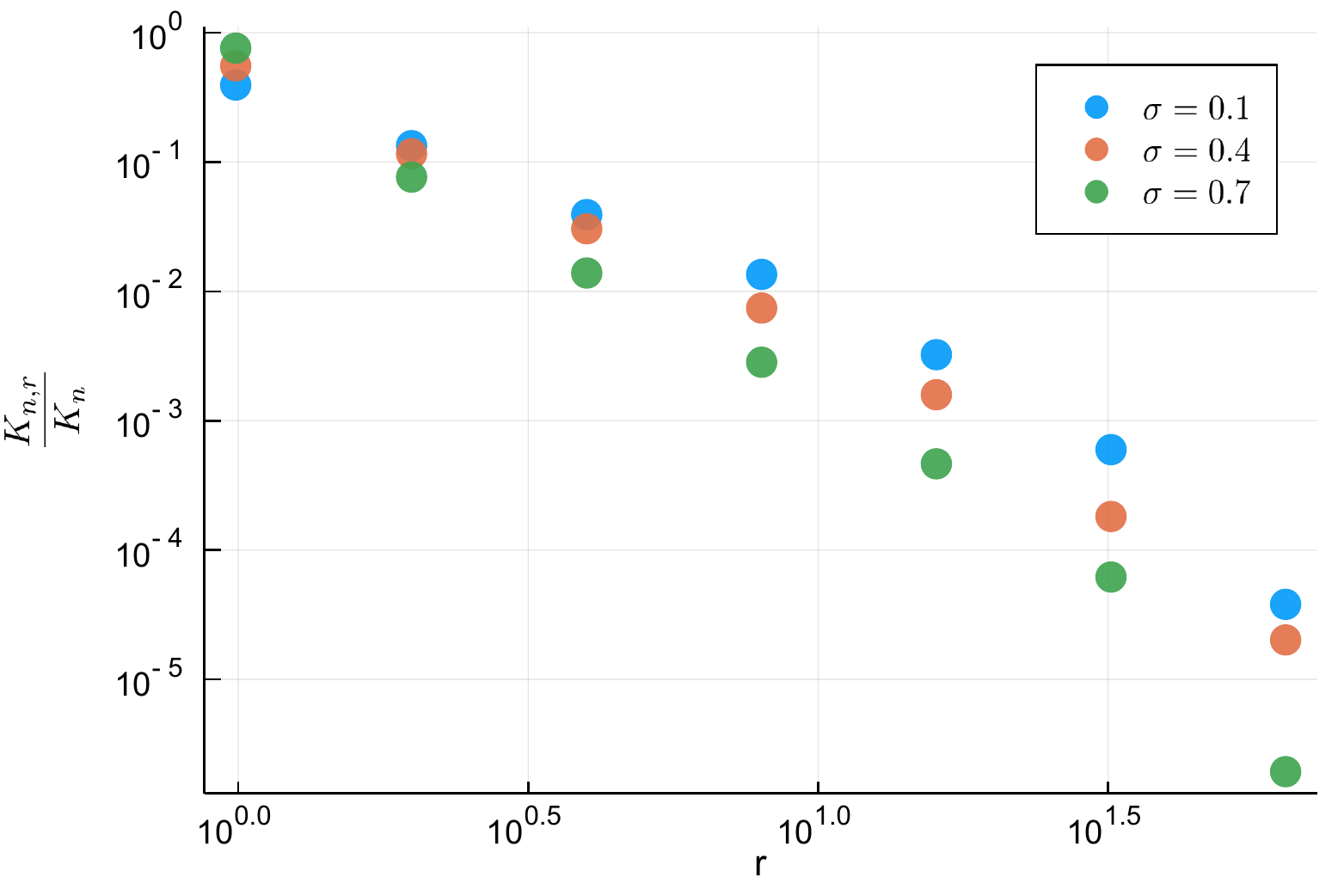}
  \caption{Log-log plot of the proportions of features shared by a fixed number of objects. Simulated examples for $\sigma\in\{0.1, 0.4, 0.7\}$, with $\xi=1$ and $\eta=30$.}
  \label{fig:powerlaw}
\end{figure}

Analogously to the three-parameter IBP model, the proposed non-exchangeable feature allocation model exhibits the power-law property for the number of features shared by a given number of objects (see Figure~\ref{fig:powerlaw} for an illustration of this property).

\begin{proposition}[power-law properties]\label{prop:powerlaw}
  Consider the model of Section \ref{model} with $\xi> 1-\sigma$ and a L\'evy measure $\rho$ whose Laplace exponent $\psi$ satisfies Equation~\eqref{eq:asymptoticspsi}. We have, almost surely as $n$ tends to infinity
  \[
  \frac{K_{n,r}}{K_n}\rightarrow\frac{\sigma\Gamma(r-\sigma)}{r!\Gamma(1-\sigma)}.
  \]
\end{proposition}

We conjecture that the condition $\xi> 1-\sigma$ introduced in the above proposition, which is needed for the proof, is not a necessary condition, and that the above proposition actually holds for any $\xi\geq 0$.

\section{Inference}\label{sec:inference}

\subsection{Data augmentation and conditional distribution}

For posterior inference, we introduce a latent process, similar to that of \cite{Caron2012BayesianNM}. For $i=1,\ldots,n$, let
$U_i =\sum_{j\ge1} u_{ij} \delta_{\theta_j}$
where $u_{ij}=1$ if $z_{ij}=0$, and
\begin{align}
u_{ij}\mid z_{ij}=1,\omega_j\sim \tExp(\Delta_i\omega_j,1)
\end{align}
$\tExp(\lambda,T)$ denotes the right-truncated exponential distribution with rate $\lambda>0$ and truncation $T>0$  with probability density function $\lambda e^{-\lambda x}(1-e^{-T\lambda})^{-1}\1{x<T}$. Note that by construction, $z_{ij}=\1{u_{ij}<1}$ is a deterministic function of $u_{ij}$.

Denote $\widetilde\theta_1,\ldots,\widetilde\theta_{K_n}$ the set of $\theta_j$'s such that $m_{n,j}>0$, $\widetilde\omega_1,\ldots,\widetilde\omega_{K_n}$ the corresponding weights, $(\widetilde u_{i,k})_{i=1,\ldots,n;k=1,\ldots,K_n}$ the corresponding latent variables, $(\widetilde z_{i,k})_{i=1,\ldots,n;k=1,\ldots,K_n}$ the corresponding binary variables and $\widetilde m_{n,1},\ldots,\widetilde m_{n,K_n}$ the associated features' sizes. For $k=1,\ldots,K_n$, let $Y^*_k=\inf_i\{Y_i:\widetilde z_{ik}=1\}$.
With analogous calculations to \cite{Caron2012BayesianNM} it is possible to write down the conditional distributions
\begin{align*}
  &\mathbb P(U_1,\dots,U_n\,|\,B)\\ &= e^{-\sum_{i=1}^n\Delta_i\left (\bar B (Y_i)-\sum_{k=1}^{K_n}\widetilde\omega_k\1{\widetilde\theta_k<Y_i}\right )} \left(\prod_{i=1}^n\Delta_i^{\sum_{k=1}^{K_n} \widetilde z_{ik}}\right) \\
  &\left(\prod_{k=1}^{K_n}\widetilde\omega_k^{\widetilde m_{n,k}}\,e^{-\widetilde\omega_k\sum_{i=1}^n\Delta_i \widetilde u_{ik}\,\1{\widetilde\theta_k<Y_i}}\1{\widetilde\theta_k<Y^*_k}\right)
\end{align*}
where $\bar B (Y_i)=B((0,Y_i))=\sum_{j}\omega_j \1{\theta_j<Y_i}$.
Using the results on Poisson partition calculus \citep{James2002}, we can marginalize out the CRM
\begin{align}
  \mathbb P(U_1,\dots,U_n) &=  e^{-\int_0^{Y_n}\psi(f_n(\theta))d\theta}\left(\prod_{i=1}^n\Delta_i^{\sum_{k=1}^{K_n} \widetilde z_{ik}}\right) \nonumber\\
  &\times\prod_{k=1}^{K_n}\kappa\left(\widetilde m_{n,k},\sum_{i=1}^n\Delta_i \widetilde u_{ik}\1{\widetilde\theta_k<Y_i}\right)\1{\widetilde\theta_k<Y^*_k}\label{eq:marginalU}
\end{align}
with $f_n(\theta):=\sum_{i=1}^n \Delta_i \1{\theta\leq Y_i}$. The conditional distribution is
\begin{align}
B\,|\, (\widetilde u_{ik}) \stackrel{d}{=} B^* + \sum_{k=1}^{K_n}\widetilde\omega_k\delta_{\widetilde\theta_k}\label{eq:conditionalB}
\end{align}
where $B^*$ is an inhomogeneous CRM  with L\'{e}vy measure $\nu^*(d\omega,d\theta) = e^{-\omega\sum_{i=1}^n\Delta_i\1{\theta<Y_i}}\rho(\omega)d\omega d\theta$, and it is independent of $(\widetilde \omega_k,\widetilde\theta_k)_{k=1,\ldots,K_n}$ which are independent (in $k$), with joint posterior probability density
\[p(\widetilde\omega_k,\widetilde\theta_k|\,(\widetilde u_{ik})) \propto \widetilde\omega_k^{\widetilde m_{n,k}} e^{-\widetilde\omega_k\sum_{i=1}^n\Delta_i \widetilde u_{ik}\1{\widetilde \theta_k<Y_i}}\rho(\widetilde\omega_k)\1{\widetilde\theta_k<Y^*_k}.\]

\subsection{Gibbs sampler}
Assume that we have observed the binary feature allocations $(\widetilde z_{i,k})_{i=1,\ldots,n,k=1,\ldots,K_n}$. We take an empirical Bayes approach, and wish to approximate the posterior distribution
\begin{equation}
p((\widetilde u_{i,k}),(\widetilde \omega_k),(\widetilde \theta_k)\mid (\widetilde z_{i,k}), \widehat\phi)
\label{eq:posterior}
\end{equation}
 where $\widehat\phi$ is a point estimate of the hyperparameters $\phi=(\xi,\sigma,\eta,\zeta)$. We explain in the next section how to obtain consistent estimators of the hyperparameters using our asymptotic results. For the GGP, a Gibbs sampler with target distribution \eqref{eq:posterior} can be derived as follows
\begin{itemize}
\item For $i=1,\ldots,n$, $k=1,\ldots,K_n$, such that $\widetilde z_{i,k}=1$ sample $\widetilde u_{ik}\mid \text{rest} \sim \tExp(\Delta_i\widetilde\omega_k ,1).$
\item For $k=1,\ldots,K_n$, sample
  \begin{small}
  \[
  \widetilde\omega_k\mid \text{rest} \sim \text{Gamma}\left (\widetilde m_{n,k}-\sigma,\zeta+\sum_{i=1}^n\Delta_i \widetilde u_{ik}\1{\widetilde\theta_k<Y_i}\right ).
  \]
  \end{small}
\item For $k=1,\ldots,K_n$, sample $$\widetilde\theta_k\mid \text{rest} \sim e^{-\widetilde\omega_k\sum_{i=1}^n\Delta_i\widetilde u_{ik}\1{\widetilde\theta_k<Y_i}}\1{\widetilde\theta_k<Y^*_k}.$$
\end{itemize}
The last distribution is piecewise constant, and one can therefore straightforwardly sample from it.

\begin{figure*}[h!]
  \centering
  \subfigure[Synthetic - NE]{\includegraphics[width=.19\textwidth]{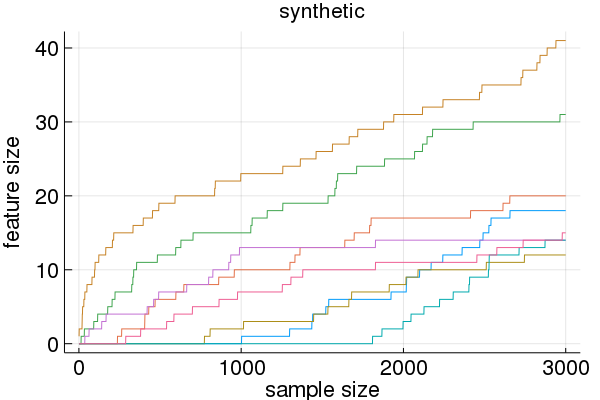}}
   \subfigure[Synthetic - IBP]{\includegraphics[width=.20\textwidth]{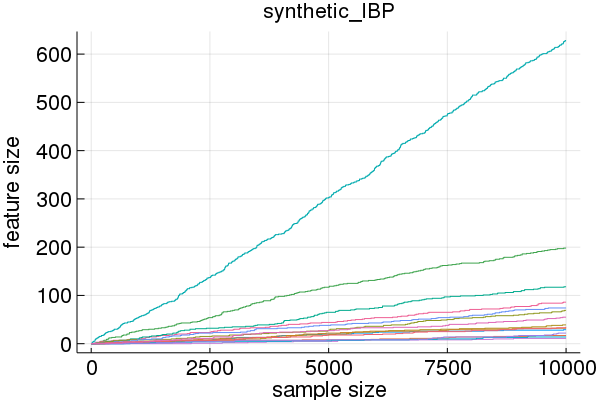}}
   \subfigure[Movies]{\includegraphics[width=.19\textwidth]{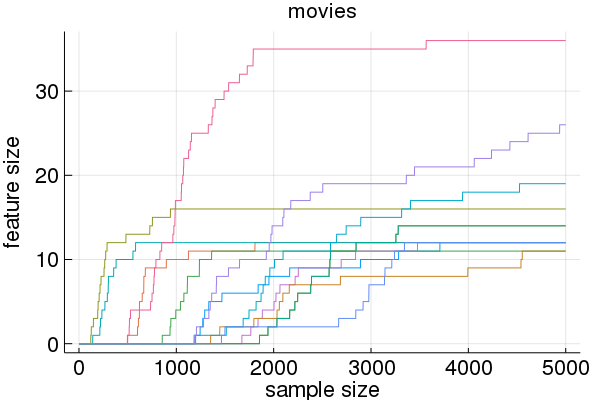}}
    \subfigure[ArXiv]{\includegraphics[width=.19\textwidth]{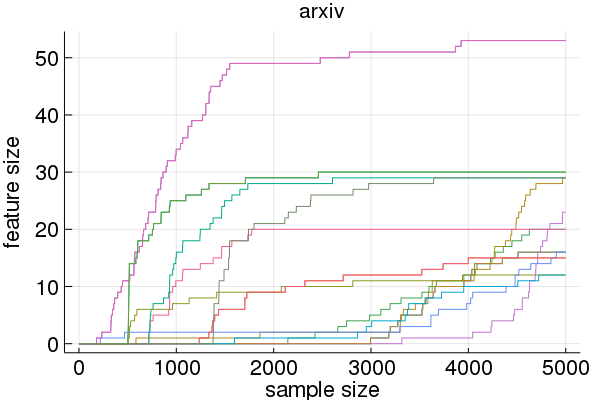}}
    \subfigure[Amazon]{\includegraphics[width=.19\textwidth]{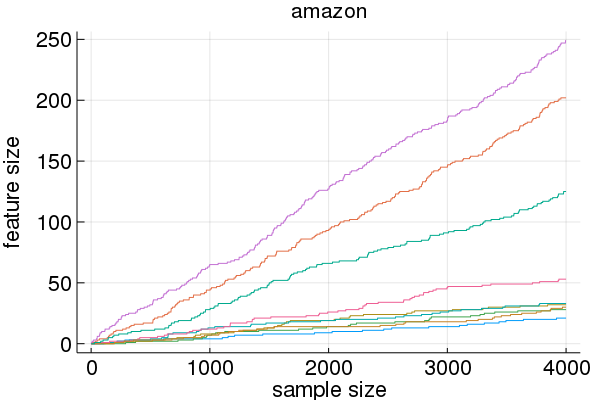}}
  \caption{Evolution of some of features' sizes with respect to the sample size for different datasets.}
  \label{fig:featuresize}
\end{figure*}

\begin{table*}[h!]
  \begin{center}
    \begin{tabular}{rllllll}
      \toprule\small
      & \multicolumn{3}{c}{Non-exchangeable} & \multicolumn{3}{c}{Indian Buffet process} \\
      & \multicolumn{1}{c}{$\hat \sigma_{\text{NE}}$} & \multicolumn{1}{c}{L2 error} & \multicolumn{1}{c}{$90\%$ CI} & \multicolumn{1}{c}{$\hat \sigma_{\text{IBP}}$} & \multicolumn{1}{c}{L2 error} & \multicolumn{1}{c}{$90\%$ CI} \\
      \midrule
      Synthetic &$0.61$ & $15.41$ & $[15.32, 15.50]$ & $0.51$ & $19.93$ & $[19.72, 20.13]$ \\
      Synthetic IBP &$0.41$& $30.47$ & $[30.37, 30.54]$ & $0.39$ & $31.19$ & $[31.11, 31.28]$ \\
      IMDb movies & $0.31$ & $21.27$ & $[21.20,21.35]$ & $0.07$ & $36.78$ & $[36.59, 37.04]$ \\
      Arxiv & $0.48$ & $7.96$ & $[7.89, 7.96]$ & $0.31$ & $13.95$ & $[13.81, 14.08]$ \\
      Amazon & $0.57$ & $142.06$ & $[141.74,142.38]$ & $0.61$ & $142.62$ & $[142.25,143.02]$ \\
      \bottomrule
    \end{tabular}
  \end{center}
    \caption{Estimated values for the parameter $\sigma$ obtained by the Indian Buffet process ($\hat \sigma_{\text{IBP}}$) and our non-exchangeable model ($\hat \sigma_{\text{NE}}$). Mean and quantiles of the $L_2$ predictive error.}
  \label{table:sigmaerr}
\end{table*}

\subsection{Estimation of the Hyperparameters}
We use here the asymptotic results of Section \ref{sec:asymptotics} to derive consistent estimators of the hyperparameters.
The parameter $\xi$ controls the features' growth and is consistently estimated using Proposition~\ref{prop:feature_size}, while Proposition~\ref{prop:powerlaw} can be used to consistently estimate the parameter $\sigma$ from the proportion of features of size one:
\[
\hat\xi = \frac{\log n}{\log \max_{k=1,\ldots,K_n} \widetilde m_{n,k}} - 1, \qquad  \hat \sigma = \frac{K_{n,1}}{K_n}.
\]
Using Propositions~\ref{prop:tot_features} and ~\ref{prop:unique_features} we have the following consistent estimators for the parameters $\eta$ and $\zeta$ (GGP):
\[
\hat \eta = \frac{\Gamma(\hat\sigma+\hat\xi +1)}{\Gamma(\hat\sigma)\Gamma(\hat\xi+1)} \frac{K_n}{n^{\frac{\hat\xi+\hat\sigma}{1+\hat\sigma}}}, \qquad \hat \zeta = \left(\frac{(\hat \xi +1)\,m_n}{\hat \eta \, n} \right)^{\frac{1}{\hat\sigma+1}}.
\]

Although these estimators are consistent, it should be noted that $\hat\xi$ depends logarithmically on the sample size, leading potentially to a slow convergence. However, as shown in the simulated experiments (see section~\ref{experiments} and Table~\ref{table:sigmaerr}), the estimated values of the hyperparamenters $\xi$ and $\sigma$ are close to the true values.

\section{Experiments}\label{experiments}

In this section we present some experiments to compare the proposed model with GG measure against the three-parameter IBP. We estimate the parameters on a training set, and compute the empirical normalized $L_2$ error between the true and predicted features' sizes on a test set
\[
\text{Err} = \frac{1}{n_{\text{test}}}\sum_{k=1}^{K_{n_{\text{train}}}}\; \sum_{i=n_{\text{train}}+1}^{n_{\text{train}}+n_{\text{test}}}
(\widetilde m^{\text{true}}_{i,k}-\widetilde m_{i,k})^2
\]
where $\widetilde m^{\text{true}}_{i,k}$ is the observed size of feature $k$, $n_{\text{train}}$ and $n_{\text{test}}$ are the numbers of objects in the training set and test set respectively and  $K_{n_{\text{train}}}$ is the number of features observed in the training set. We aim at reporting the posterior mean $L_2$ error $\mathbb E[\text{Err}\mid (\widetilde z_{ik})_{i=1,\ldots,n_{\text{train}},k=1,\ldots,K_{n_{\text{train}}}} ]$  and the 90\% credible interval of the posterior distribution of the $L_2$ error. Details on the posterior predictive under our non-exchangeable model are given in Section \ref{sec:posteriorpredictive} of the Supplementary Material. In what follows we denote the total number of objects in the dataset as $n$. % = n_{\text{train}} + n_{\text{test}}$.
Given the closed form expression~\eqref{eq:featureallocationIBP}, the hyperparameters $(\eta, \sigma, \zeta)$ of the three-parameter IBP are estimated by maximum likelihood.

\begin{figure*}[t]
  \centering
  \includegraphics[width=.19\textwidth]{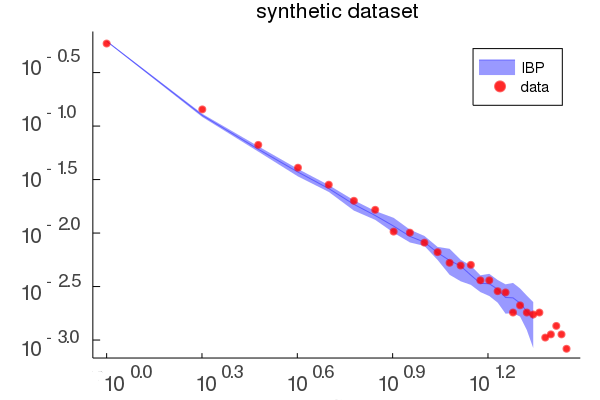}
  \includegraphics[width=.19\textwidth]{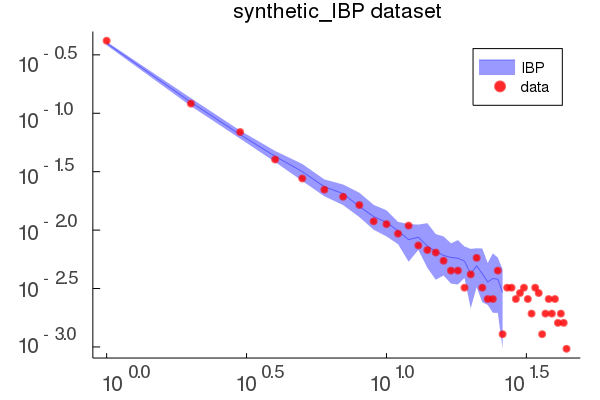}
  \includegraphics[width=.19\textwidth]{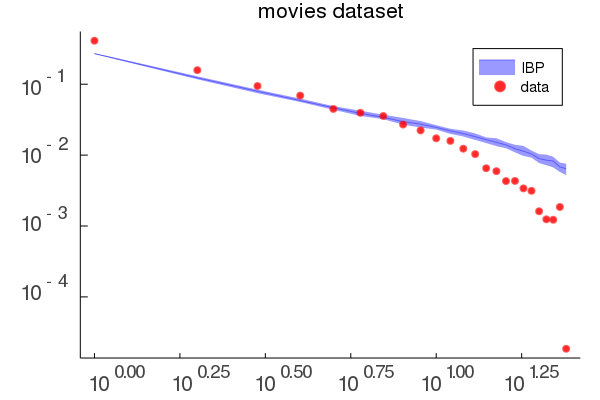}
  \includegraphics[width=.19\textwidth]{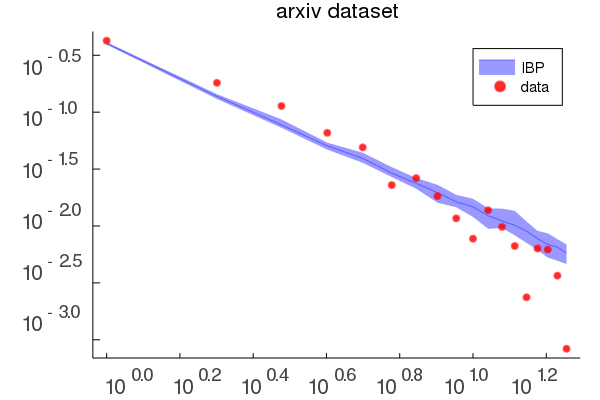}
  \includegraphics[width=.19\textwidth]{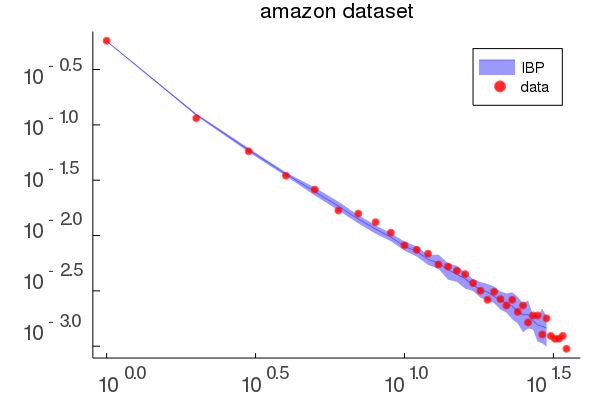}\\
  \subfigure[Synthetic - NE]{\includegraphics[width=.19\textwidth]{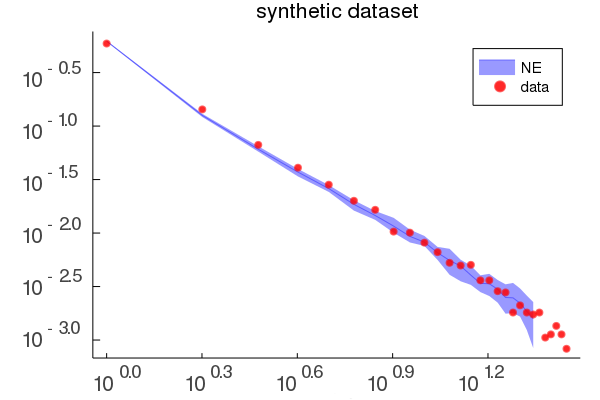}}
  \subfigure[Synthetic - IBP]{\includegraphics[width=.20\textwidth]{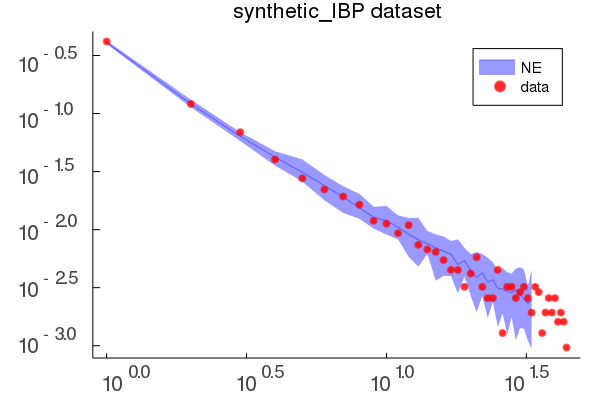}}
  \subfigure[Movies]{\includegraphics[width=.19\textwidth]{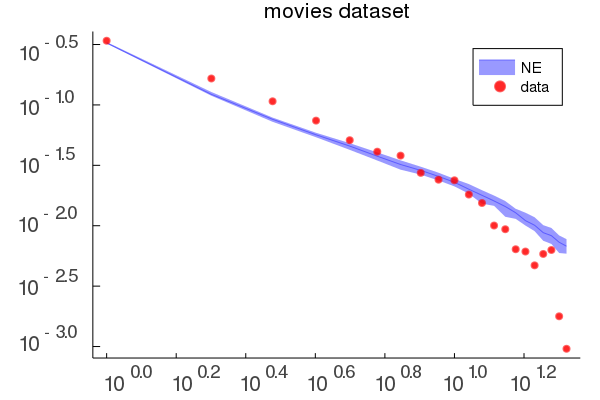}}
  \subfigure[ArXiv]{\includegraphics[width=.19\textwidth]{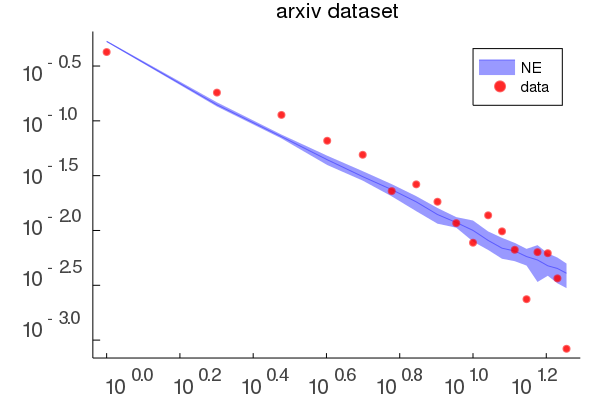}}
  \subfigure[Amazon]{\includegraphics[width=.19\textwidth]{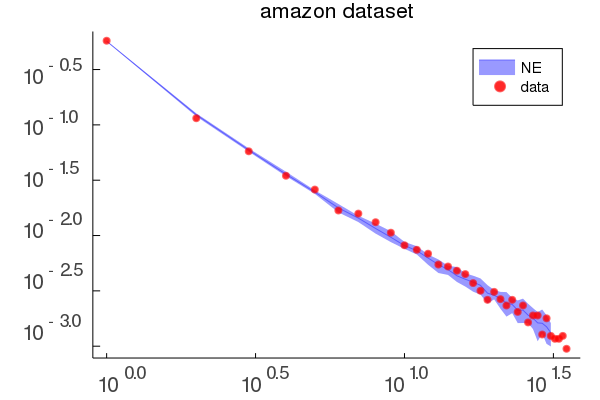}}
  \caption{Log-log plots of the empirical proportions $(K_{n,r}/K_n)_{r\ge 1}$
    %of features shared by a given number of objects
    (red dots) and $95\%$ posterior predictive intervals (blue). Top: Indian Buffet Process (IBP); bottom: non-exchangeable model (NE).}
  \label{fig:freqplots}
\end{figure*}

\begin{figure}[t]
  \centering
  \includegraphics[height=4.3cm]{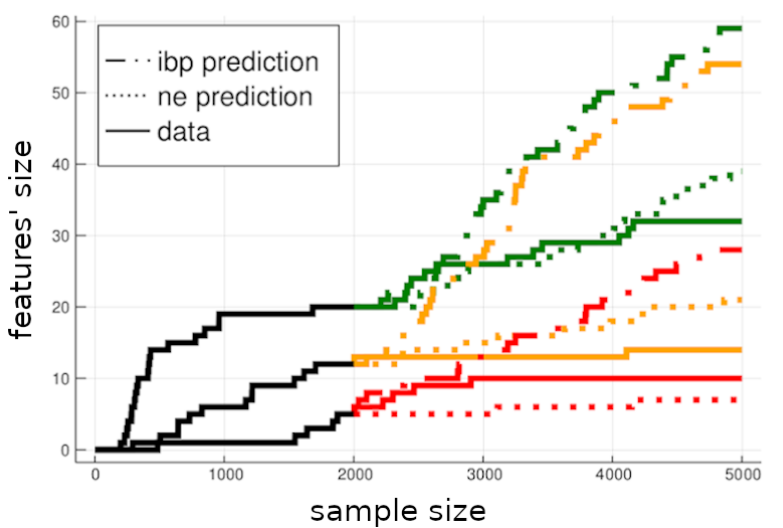}
  \caption{Movies dataset: sizes of some of the features (solid lines, training set in black), predicted features' sizes by the non-exchangeable model (dot lines), predictions by the IBP (dash-dot lines).}
  \label{fig:feature_prediction}
\end{figure}

\textbf{Synthetic data.} The models are first tested on synthetic datasets. We generate a binary matrix with $n=3000$ rows from our non-exchangeable model with parameters $(\eta,\sigma,\xi,\zeta) = (30, 0.5, 1, 1)$, with a training set of size $n_\text{train}=1000$. Figure~\ref{fig:featuresize} shows that the feature' sizes are clearly sublinear, and as expected the prediction error is higher under the IBP model, as shown in Table~\ref{table:sigmaerr}. The second synthetic dataset of size $n=10000$ is generated from the IBP model with parameters $(\eta,\sigma,\zeta)=(20, 0.4, 2.4)$, with $n_\text{train}=3000$. The estimated value for the parameter $\xi$ in the non-exchangeable model is close to zero ($\hat \xi = 0.04$). This shows that our model is able to capture the linearity in the feature size, typical of the exchangeable datasets. Both models are able to recover the parameter $\sigma$ correctly and the prediction errors are about the same (see Table~\ref{table:sigmaerr} and Figure~\ref{fig:featuresize}).

\textbf{Amazon data.} This real dataset contains time-ordered reviews of movies from Amazon\footnote{https://snap.stanford.edu}. We consider the first $n=4000$ reviews and use $n_{\text{train}}=1000$ reviews for training. In this context the binary matrix represents the presence of words in the reviews. Figure~\ref{fig:featuresize} shows that the presence of words in the reviews increases linearly with the number of reviews, which is expected to be observed in most of text data. As a consequence, the estimated parameter $\hat\xi = 0.01$ is very close to zero and the prediction errors of the two models are the same (see Table~\ref{table:sigmaerr} and Figure~\ref{fig:featuresize}).

\textbf{ArXiv data.} The ArXiv dataset\footnote{https://www.kaggle.com/neelshah18/arxivdataset} contains time-ordered articles uploaded on the Arxiv website. Each article is represented as a binary array that encodes its authors. Features size then represents the number of articles published by an author as a function of all the articles published on ArXiv. It is therefore expected a sublinear trend of the features' sizes, which is shown in Figure~\ref{fig:featuresize}. We considered the first $n=5000$ movies and used the first $n_{train}=1000$ for training. The non-exchangeable model captures this asymptotic property ($\hat\xi = 0.96$) resulting in a better predictive performance compared to the IBP (see Table~\ref{table:sigmaerr} and Figure~\ref{fig:featuresize}).

\textbf{Movies data.} This dataset\footnote{http://udbms.cs.helsinki.fi/?datasets/film\_dataset} contains time-ordered movies listed in Wikipedia and IMDB. Here the features are represented by the actors who have performed in the movie. The number of movies in which a specific actor has performed is obviously sublinear. We consider a subset of $n=5000$ movies with $n_{\text{train}}=2000$. The sublinear trend of the features' sizes is correctly modelled by our model ($\hat\xi = 0.45$) which outperforms the IBP in the prediction error (see Table~\ref{table:sigmaerr} and Figure~\ref{fig:featuresize} and ~\ref{fig:feature_prediction}).

Finally, Figure~\ref{fig:freqplots} shows the posterior predictive of the proportions of features shared by a given number of objects. Both models have the same asymptotic power-law property for this statistic and their fit is similar across all the datasets as expected.

\section{Discussion}
\label{sec:discussion}
Various feature allocation models have been proposed in the literature that relax the exchangeability assumption. In particular dependent Indian buffet processes~\citep{Caron2009,Williamson2010a,Zhou2011,Ren2011,Miller2012,Gershman2014,Perrone2017} include models with covariates (time, space, graph, etc.), so that objects with similar covariates have similar feature weights; see \citep{Foti2013} for a review. Contrary to this line of work, the aim here is to derive feature allocation models with provable sublinear growth of the features' sizes, retaining the good asymptotic properties of the exchangeable models, namely the power-law behaviour and the control on the number of unique features. The proposed class of models allows to control these quantities by interpretable and tunable parameters, and inference is carried out by a simple Gibbs algorithm.
%Our construction shares some similarities with the model of \cite{Caron2012BayesianNM} for bipartite graphs, and we use in Section \ref{sec:inference} a similar data augmentation scheme for efficient posterior inference. 
Finally, the problem addressed in this paper is closely related to the problem of microclustering~\citep{Miller2015}, which aims at finding models for random partitions where the size of the clusters grows sublinearly.

\paragraph{Acknowledgments.} FC acknowledges support from EPSRC under grant EP/P026753/1 and from the Alan Turing Institute under EPSRC grant EP/N510129/1. GDB is funded by EPSRC under grant EP/L016710/1.

\newpage

%\nocite{*}
%\bibliographystyle{apalike}
\bibliography{ref_features}
\clearpage
%\appendix

%\newpage

{\large \textbf {Supplementary Material} }
\vspace{0.5cm}

In this document we provide the proofs of the Propositions stated in the main paper, together with the posterior predictive distribution of the feature allocations.

\section{Posterior predictive distribution of feature allocations}
\label{sec:posteriorpredictive}
The data augmentation described in Section \ref{sec:inference}  in the main paper allows us to compute the posterior predictive distribution of the $n+1$-th object given the first $n$ ones.
\begin{align*}
  Z_{n+1}\,|\,U_1,\dots,U_n \stackrel{d}{=}Z^*_{n+1}+\sum_{k=1}^{K_n}\widetilde z_{n+1,k}\delta_{\widetilde\theta_k}
\end{align*}
where $Z^*_{n+1}$ is independent of the $\widetilde z_{n+1,k}$ which are distributed as follows
\begin{align*}
  &\widetilde z_{n+1,k}\,|\,U_1,\dots,U_n\sim\\
  &\text{Ber}\left(1-\frac{\kappa\left (\widetilde m_{nj}\,,\,\Delta_{n+1}+\sum_{i=1}^n\Delta_i\widetilde u_{ij}\1{\widetilde \theta_k\leq Y_i}\right )}{\kappa\left (\widetilde m_{nj}\,,\,\sum_{i=1}^n\Delta_i\widetilde u_{ij}\1{\widetilde \theta_k\leq Y_i}\right)}\right).
\end{align*}

By the marking theorem for Poisson point processes and Equation \eqref{eq:conditionalB}, $Z^*_{n+1}=\sum_{k=K_n+1}^{K_n+K_{n+1}^* }\delta_{\widetilde \theta_{k}}$ is a Poisson random measure on $\mathbb R_+$ with mean measure  $\mu_{n+1}^*(d\theta)=\int_0^\infty \left(1-e^{-\omega\Delta_{n+1}\1{\theta\leq Y_{n+1}}}\right)\,d\nu^*(d\omega,d\theta)$, where we recall that
\begin{align*}
\nu^*(d\omega,d\theta) &= e^{-\omega\sum_{i=1}^n\Delta_i\1{\theta\leq Y_i}}\rho(\omega)d\omega d\theta\\
&=\left (\1{\theta>Y_n}+\sum_{i=1}^n e^{-\omega(T_n-T_{i-1})}\1{Y_{i-1}<\theta\leq Y_i}\right )\\&~~\times\rho(\omega)d\omega d\theta
\end{align*}
where $T_0=Y_0=0$. Therefore, the number $K_{n+1}^*=Z^*_{n+1}(\mathbb R_+)$ of new features of the $n+1$ object is Poisson distributed with mean
\begin{align*}
  &\mathbb E [Z^*_{n+1}(\mathbb R_+)]\\&=\int_0^\infty\int_0^\infty \left(1-e^{-\omega\Delta_{n+1}\1{\theta\leq Y_{n+1}}}\right)\,d\nu^*(d\omega,d\theta)
\end{align*}
If follows that
\begin{align}
  &\mathbb E [Z^*_{n+1}(\mathbb R_+)]\nonumber\\
  & =\sum_{i=1}^{n+1}(Y_{i}-Y_{i-1}) \int_0^\infty \left(1-e^{-\omega\Delta_{n+1}}\right)e^{-\omega(T_n-T_{i-1})}\,\rho(\omega)d\omega\label{eq:temp2}
\end{align}

The locations $\widetilde\theta_{K_n+1},\ldots,\widetilde \theta_{K_n+K^*_{n+1}}$ are sampled iid from the piecewise constant distribution on $[0,Y_{n+1}]$ with pdf proportional to
\begin{equation}
\sum_{i=1}^{n+1}\1{Y_{i-1}<\theta\leq Y_i} \int_0^\infty \left(1-e^{-\omega\Delta_{n+1}}\right)e^{-\omega(T_n-T_{i-1})}\,\rho(\omega)d\omega\label{eq:temp3}
\end{equation}

If $B$ is a GGP, the integral in Equations~\eqref{eq:temp2} and \eqref{eq:temp3} is tractable and we have
\begin{align*}
&\int_0^\infty \left(1-e^{-\omega\Delta_{n+1}}\right)e^{-\omega(T_n-T_{i-1})}\,\rho(\omega)d\omega\\
&=\left\{
    \begin{array}{ll}
      \frac{\eta}{\sigma}\left[(T_{n+1}-T_{i-1}+\zeta)^\sigma -(T_{n}-T_{i-1}+\zeta)^\sigma\right] & \sigma>0 \\
      \eta\log\left(1+\frac{\Delta_{n+1}}{T_n-T_{i-1}+\zeta}\right) & \sigma=0 \\
    \end{array}
  \right.
\end{align*}

\begin{proof}
We have
$$
\Pr(\widetilde z_{n+1,k}=1\,|\widetilde \omega_k,\,U_1,\dots,U_n)=1-e^{-\widetilde \omega_k\Delta_{n+1}}
$$
and
$$
p(\widetilde \omega_k\mid U_1,\dots,U_n)=\frac{\widetilde \omega_k^{\widetilde m_{n,k}}e^{-\widetilde \omega_k\sum_{i=1}^n \Delta_i\widetilde u_{ij} \1{\widetilde \theta_k\leq Y_i}}\rho(\widetilde\omega_k)}{\kappa\left (\widetilde m_{n,k},\sum_{i=1}^n \Delta_i\widetilde u_{ij} \1{\widetilde \theta_k\leq Y_i} \right )}
$$
Hence
\begin{align*}
&\Pr(\widetilde z_{n+1,k}=1\,\mid \,U_1,\dots,U_n)\\&=1-\frac{\kappa\left (\widetilde m_{n,k},\Delta_{n+1}+\sum_{i=1}^n \Delta_i\widetilde u_{ij} \1{\widetilde \theta_k\leq Y_i} \right )}{\kappa\left (\widetilde m_{n,k},\sum_{i=1}^n \Delta_i\widetilde u_{ij} \1{\widetilde \theta_k\leq Y_i} \right )}
\end{align*}
\end{proof}

The conditional distribution of the latent point process $U_{n+1}$ can be written as follows:
\begin{align*}
  &p(\widetilde u_{n+1,k}\,|\, \widetilde z_{n+1,k}=1, \, \widetilde u_{1:n,k}) \\ &\propto \kappa\left(\widetilde m_{n,k}+1,\Delta_{n+1}\widetilde u_{n+1,k} +
  \sum_{i=1}^n\Delta_i \widetilde u_{ik}\right) \1{u_{n+1,j}<1}
\end{align*}

\begin{proof}
We have
\begin{align*}
&p(\widetilde u_{n+1,k}\,|\,\widetilde\omega_k, \widetilde z_{n+1,k}=1, \, \widetilde u_{1:n,k})\\&=\frac{\Delta_{n+1}\widetilde\omega_k e^{-\widetilde u_{n+1,k}\Delta_{n+1}\widetilde\omega_k} \1{\widetilde u_{n+1,k}<1}}{1-e^{-\Delta_{n+1}\widetilde\omega_k}}
\end{align*}
and
\begin{align*}
&p(\widetilde \omega_k\mid \widetilde z_{n+1,k}=1,  U_1,\dots,U_n)\\&\propto (1-e^{-\Delta_{n+1}\widetilde\omega_k})\widetilde \omega_k^{\widetilde m_{n,k}}e^{-\widetilde \omega_k\sum_{i=1}^n \Delta_i\widetilde u_{ij} \1{\widetilde \theta_k\leq Y_i}}\rho(\widetilde\omega_k)
\end{align*}
\end{proof}

\section{Proofs}

\subsection{Proof of Proposition \ref{prop:nbfeatureobject}}

By the marking theorem for Poisson point processes, the set of points $\{(\omega_j)_{j\geq 1}\mid z_{ij}=1\}$ is drawn from a Poisson point process with mean measure $Y_i (1-e^{-\Delta_i\omega})\rho(\omega)d\omega$. The total number of such points $Z_i(\mathbb R_+)$ is therefore Poisson distributed with mean $\mathbb E \left [Z_i(\mathbb R_+)\right ]=Y_i\psi(\Delta_i)$. Using integration by part, we have
$$
\psi(t)=t\int_0^\infty e^{-wt}\overline\rho(w)dw
$$
where
\begin{equation}
\overline\rho(x)=\int_x^\infty \rho(w)dw.
\end{equation}
Hence, by monotone convergence,
$$
\lim_{t\rightarrow\infty}\frac{\psi(t)}{t}=\int_0^\infty \overline\rho(w)dw=\kappa(1,0)
$$
and it follows that $$Y_n\psi(\Delta_n)\obsim{n\to\infty}Y_n\Delta_n\kappa(1,0)$$
Finally note that $\Delta_n\obsim{n\to\infty}(1+\xi)^{-1}n^{-\xi/(\xi+1)}$, hence $Y_n\Delta_n\rightarrow (1+\xi)^{-1}$.

\subsection{Proof of Proposition \ref{prop:tot_features}}

  The number of features observed in the first $n$ objects can be written as
\[m_n := \sum_{i=1}^n\sum_{j\ge1} z_{ij} = \sum_{i=1}^n Z_i(\mathbb R_+)\]
Since $\mathbb E[Z_n(\mathbb R_+)] = \mathbb E[m_n]-\mathbb E[m_{n-1}]$, it follows by Stolz-Ces\`{a}ro theorem  that
\[\mathbb E [m_n]\obsim{n\to\infty} (1+\xi)^{-1} \kappa(1,0) n.\]
In order to get the almost sure convergence of $m_n$ to its expectation we can use the Kolmogorov strong law of large numbers which, under the assumption $\sum_{n\ge 1}\text{Var}\left(\frac{Z_n(\mathbb R_+)}{n}\right)<\infty$, gives
\[
\frac{m_n - \mathbb E[m_n]}{n} \rightarrow 0 \text{ almost surely.}
\]
Recall that $\text{Var}(Z_n(\mathbb R_+))=\mathbb E[Z_n(\mathbb R_+)]$.
Therefore the summability condition on the variance boils down to the convergence of the sum $\sum_{n\ge 1} \frac{1}{n^2}Y_n\,\psi(\Delta_n)$, which holds true since the elements of the sum are of order $ n^{-2}$.

\subsection{Proof of Proposition \ref{prop:feature_size}}

  Since $\mathbb E [m_{n,j}|B] = \sum_{i=1}^n \left(1-e^{-\omega_j\Delta_i}\right)\1{\theta_j<Y_i}$, we have
 $\frac{\mathbb E [m_{nj}|B] - \mathbb E [m_{n-1\, j}|B]}{T_n -T_{n-1}}= \frac{1-e^{-\omega_j\Delta_n}}{\Delta_n}\1{\theta_j<Y_n}\rightarrow \omega_j$, then by Stolz-Ces\`{a}ro theorem we have that
  \[\mathbb E [m_{nj}|B]\obsim{n\to\infty} \omega_j T_n.\]
  We have
  \begin{align*}
    \text{Var}(m_{n,j}\,|\,B) &=\sum_{i=1}^n \left(1-e^{-\omega_j\Delta_i}\right)e^{-\omega_j\Delta_i}\1{\theta_j<Y_i}\\ &\leq \mathbb E [m_{n,j}|B].
    \end{align*}
  Using the sandwiching argument in Proposition 2 of \cite{Gnedin2007} it follows that, conditionally on $B$, $\frac{m_{nj}}{\mathbb E[m_{nj}\,|\,B]} \to 1$ almost surely.

\subsection{Proof of Proposition \ref{prop:unique_features}}

Applying Campbell's theorem
\begin{align*}
\mathbb E[K_n] &=\mathbb E[\mathbb E[K_n\mid B]]\\
&=\mathbb E\left [\sum_j \Pr(m_{n,j}>0\mid B)\right ]\\
 &= \int_0^\infty\int_0^\infty \left(1-e^{-\omega f_n(\theta)}\right)\rho(\omega)d\omega d\theta\\
&=\int_0^{Y_n} \psi(T_n-g(\theta))d\theta
\end{align*}
where
\begin{align}
f_n(\theta):=\sum_{i=1}^n \Delta_i \1{\theta\leq Y_i} = (T_n - g(\theta))\1{\theta\leq Y_n}\label{eq:fn}
\end{align}
 with $g(\theta):= \sum_{i=1}^\infty \Delta_i \1{\theta>Y_i}$ is a monotone increasing step function satisfying, for all $\theta\geq 0$
\begin{align}
\max(0,g_1(\theta)) \leq g(\theta)\leq g_2(\theta)\label{eq:ineqg}
\end{align}
where $g_1(\theta)=(\theta^{(\xi+1)/\xi}-1)^{1/(\xi+1)}$ and $g_2(\theta)= \theta^{1/\xi}$. Note that $g_1(\theta)\obsim{\theta\to\infty}g_2(\theta)\obsim{\theta\to\infty}\theta^{1/\xi}$. As $\psi$ is an increasing function, it follows
\begin{align*}
\int_0^{Y_n} \psi(T_n-g_2(\theta))d\theta\leq\mathbb E[K_n]\leq & \int_1^{Y_n} \psi(T_n-g_1(\theta))d\theta\\&~+\psi(T_n).
\end{align*}
Using a change of variable, we obtain
\begin{align*}
\int_0^{Y_n}\psi(T_n-g_2(\theta))d\theta =\xi \int_0^{T_n}\psi(T_n-\theta)\,\theta^{\xi-1}d\theta
\end{align*}
Finally, noting that
$$
\psi(t)\obsim{t\to\infty}t^\sigma \ell(t)
$$
where
$$
\ell(t)=\left \{
\begin{array}{ll}
  \eta\log(t) & \sigma=0 \\
  \frac{\eta}{\sigma}  & \sigma\in(0,1)
\end{array} \right .
$$

and using \cite[Lemma 14]{DiBenedetto2017}, we obtain
\begin{align*}
\int_0^{Y_n}\psi(T_n-g_2(\theta))d\theta\obsim{n\to\infty} \frac{\Gamma(\xi+1)\Gamma(\sigma+1)}{\Gamma(\sigma+\xi+1)}\,n^{\frac{\xi+\sigma}{\xi+1}}\ell(n)
\end{align*}
Similarly, we have
\begin{align*}
\int_1^{Y_n}\psi(T_n-g_1(\theta))d\theta\obsim{n\to\infty} \frac{\Gamma(\xi+1)\Gamma(\sigma+1)}{\Gamma(\sigma+\xi+1)}\,n^{\frac{\xi+\sigma}{\xi+1}}\ell(n).
\end{align*}
It follows by sandwiching that
\begin{align*}
\mathbb E[K_n]\obsim{n\to\infty} \frac{\Gamma(\xi+1)\Gamma(\sigma+1)}{\Gamma(\sigma+\xi+1)}\,n^{\frac{\xi+\sigma}{\xi+1}}\ell(n).
\end{align*}
Using Campbell's theorem again,
\begin{align*}
&\text{Var}[K_n]=\text{Var}[\mathbb E[K_n\mid B]]+\mathbb E[\text{Var}[K_n\mid B]] \\&= \int \left(1-e^{-\omega f_n(\theta)}\right)e^{-\omega f_n(\theta)}\rho(d\omega)d\omega d\theta \\&~~+ \int \left(1-e^{-\omega f_n(\theta)}\right)^2\rho(d\omega)d\omega d\theta
\\ &= \int \left(1-e^{-\omega f_n(\theta)}\right)\rho(d\omega)d\omega d\theta
\end{align*}
therefore the almost sure asymptotic equivalence follows by Chebyshev inequality and the strong law of large numbers for $K_n$ (see \cite[Proposition 2]{Gnedin2007}).

\subsection{Proof of Proposition \ref{prop:powerlaw}}

We have the following inequality, for any $x\geq 0$
\begin{align}
0\leq x-(1-e^{-x})\leq \frac{x^2}{2}.\label{eq:expinequality}
\end{align}

  Let us recall that
  \[ K_{n,r} = \sum_{j\ge1}\1{m_{n,j}=r}. \]
  where $m_{n,j}=\sum_{i=1}^nz_{ij}$.
  Conditional on the CRM $B$ we have
  \[ \mathbb E [K_{n,r}\,|\, B] = \sum_{j\ge1}\Pr\left(m_{n,j}=r\,\Big|\, B\right). \]
  Note that $\Pr\left(m_{n,j}=r\,\Big|\, B\right)$ only depends on $(\theta_j,\omega_j)$. Write $S_{n,r}(\theta_j,\omega_j)=\Pr\left(m_{n,j}=r\,\Big|\, B\right)$.
  Let us denote  $q_{i}(\theta_j,\omega_j):=\Pr(z_{ij}=1\mid B)=1-e^{-\Delta_i \omega_j\1{\theta_j\leq Y_i}}$; $\; \lambda_{n}(\theta_j,\omega_j):=\sum_{i=1}^nq_{i}(\theta_j,\omega_j)$. Conditional on $B$ the random variable $m_{n,j}$ has a Poisson-Binomial distribution with parameters $(q_{1}(\theta_j,\omega_j),\dots,q_{n}(\theta_j,\omega_j))$. For each fixed $(\omega_j,\theta_j)$ Le Cam's inequality~\cite{LeCam1960} and inequality \eqref{eq:expinequality} give
  \begin{align}
  &\sum_{r\ge 0}\Bigg | \Pr(m_{n,j}=r\mid B)-\Poisson\left (r;\lambda_{n}(\theta_j,\omega_j)\right ) \Bigg |\nonumber\\
  &\le 2 \sum_{i=1}^n q_{i}(\theta_j,\omega_j)^2\leq 2\omega^2_j \sum_{i=1}^n \Delta_i^2 \1{\theta_j\leq Y_i} .\label{eq:boundlecam}
  \end{align}
  where $\Poisson(r;\lambda)$ denote the probability mass function of a Poisson random variable with rate parameter $\lambda$ evaluated at $r$.
  Note that for any $0<\lambda_1\leq \lambda_2$, using coupling inequalities (see. e.g. \cite[Example 4.10 p. 154]{Roch2015})
  \begin{align*}
  \sum_{r\geq 0}\Bigg |\Poisson(r;\lambda_1)-\Poisson(r;\lambda_2)\Bigg |\leq 2(\lambda_2-\lambda_1).
  \end{align*}
  Noting that $\lambda_{n}(\theta_j,\omega_j)\leq \omega_j f_n(\theta_j)$, where $f_n$ is defined in Equation \eqref{eq:fn}, and using inequality \eqref{eq:expinequality}, we obtain
  \begin{align*}
  &\sum_{r\geq 0}\Bigg |\Poisson(r;\lambda_{n}(\theta_j,\omega_j))-\Poisson(r;\omega_j f_n(\theta_j))\Bigg |\\
  &\leq 2\sum_{i=1}^n \left (\omega_j \Delta_i- (1-e^{-\omega_j\Delta_j})\right )\1{\theta_j\leq Y_i} \\
  &\leq \omega_j^2\sum_{i=1}^n \Delta^2_i\1{\theta_j\leq Y_i}
  \end{align*}
  Combining the above inequality with the inequality \eqref{eq:boundlecam}, we obtain the total variation bound
   \begin{align}
  &\sum_{r\ge 0}\Bigg | \Pr(m_{n,j}=r\mid B)-\Poisson\left (r;\omega_j f_{n}(\theta_j)\right ) \Bigg |\nonumber\\&\qquad\leq 3\omega^2_j \sum_{i=1}^n \Delta_i^2 \1{\theta_j\leq Y_i} .\label{eq:boundmnj}
  \end{align}

  Using Campbell's theorem,
  \begin{align}
 \mathbb E\left [\sum_{j\geq 1} \omega_j^2\sum_{i=1}^n \Delta^2_i\1{\theta_j\leq Y_i} \right ]&=\kappa(2,0)\sum_{i=1}^n Y_i\Delta_i^2\nonumber\\
 &\obasymp{n\to\infty} n^{1/(1+\xi)}. \label{eq:asymperror}
  \end{align}

  Using Campbell's theorem again,
  \begin{align*}
 &\mathbb E\left [\sum_{j\geq 1}\Poisson\left (r;\omega_j f_{n}(\theta_j)\right ) \right ]\\
 &=\frac{1}{r!}\int_0^\infty\int_0^\infty e^{-\omega f_n(\theta)}\omega^r f_n(\theta)^r \,\rho(\omega)\,d\omega d\theta\\
 &=\frac{1}{r!}\int_0^\infty \kappa(r,f_n(\theta) ) f_n(\theta)^r  d\theta\\
 &=\frac{1}{r!}\int_0^{Y_n} \kappa(r,T_n-g(\theta) ) (T_n-g(\theta)))^r  d\theta\\
   \end{align*}
   We use again the inequality \eqref{eq:ineqg} to bound the above expression. The upper bound is given by
   \begin{align*}
   \frac{1}{r!}\int_0^{Y_n} \kappa(r,T_n-g_2(\theta) ) (T_n-g_1(\theta))^r  d\theta
   \end{align*}
  Using a change of variable, we obtain
  \begin{align}
   &\frac{\xi}{r!}\int_0^{T_n} \kappa(r,T_n-\theta ) (T_n-g_1(\theta^\xi)))^r \theta^{\xi-1}  d\theta\nonumber\\
   &=\frac{\xi}{r!}\int_0^{T_n} \kappa(r,\theta ) (g_1(\theta^\xi))^r (T_n-\theta)^{\xi-1}  d\theta.\label{eq:temp1}
   \end{align}
   Noting that $\kappa(r,\theta )\obsim{\theta\to\infty} \eta\theta^{\sigma-r}\frac{\Gamma(r-\sigma)}{\Gamma(1-\sigma)}$ and $(g_1(\theta^\xi))^r\obsim{\theta\to\infty} \theta^r$, and using \cite[Lemma 14]{DiBenedetto2017}, we obtain that \eqref{eq:temp1} is asymptotically equivalent to
  $$
  \eta\frac{\Gamma(\xi+1)\Gamma(r-\sigma)}{r!\Gamma(\sigma+\xi+1)}\,n^{\frac{\xi+\sigma}{\xi+1}}.
  $$
  A similar asymptotic equivalence is obtained for the lower bound, and we conclude by sandwiching that
  \begin{align*}
  &\mathbb E\left [\sum_{j\geq 1}\Poisson\left (r;\omega_j f_{n}(\theta_j)\right ) \right ] \\&~~~\obsim{n\to\infty} \eta\frac{\Gamma(\xi+1)\Gamma(r-\sigma)}{r!\Gamma(\sigma+\xi+1)}\,n^{\frac{\xi+\sigma}{\xi+1}}
  \end{align*}
  Combining the above asymptotic result with Equations \eqref{eq:boundmnj} and \eqref{eq:asymperror}, and assuming $\xi+\sigma>1$, we conclude
  \begin{align*}
  \mathbb E [K_{n,r}]& =\mathbb E \left[ \sum_{j\ge1}\Pr\left(m_{n,j}=r\,\Big|\, B\right)\right ]\\
  &\obsim{n\to\infty} \eta\frac{\Gamma(\xi+1)\Gamma(r-\sigma)}{r!\Gamma(\sigma+\xi+1)}\,n^{\frac{\xi+\sigma}{\xi+1}} .
  \end{align*}

  The variance of $K_{n,r}$ can be written as
  \begin{align*}
    \text{Var}[K_{n,r}]& =\text{Var}[\mathbb{E}[K_{n,r}\,|\,B]]+ \mathbb{E}[\text{Var}[K_{n,r}\,|\,B]]\\
    &=\int S_{n,r}(\theta,\omega)\rho(\omega)d\omega d\theta\\
    &+ \int  S_{n,r}(\theta,\omega)(1-S_{n,r}(\theta,\omega))\,\rho(\omega)d\omega d\theta \\
    &=\mathbb E[K_{n,r}].
  \end{align*}
  Using the result below Proposition 2 in \cite{Gnedin2007}, we obtain, almost surely, $\mathbb E \left [\sum_{r\geq j}K_{n,r}\right ]\obsim{n\to\infty} \sum_{r\geq j}K_{n,r}$. Using a proof similar to that of Corollary 21 in \cite{Gnedin2007}, we obtain
  \begin{align}
  K_{n,r}\obsim{n\to\infty} \eta\frac{\Gamma(\xi+1)\Gamma(r-\sigma)}{r!\Gamma(\sigma+\xi+1)}\,n^{\frac{\xi+\sigma}{\xi+1}} .\label{eq:asymptoticsKnr}
  \end{align}
  Combining Equation \eqref{eq:asymptoticsKnr} with Equation~\eqref{eq:asymptoticsKn} gives the final result.

\end{document}